\documentclass[letterpaper, 10 pt, conference]{ieeeconf}  

\IEEEoverridecommandlockouts                              

\usepackage{graphics} 
\usepackage{epsfig} 
\usepackage{mathptmx} 
\usepackage{times} 
\usepackage{amsmath} 
\usepackage{amssymb}  
\usepackage{todonotes}
\usepackage{etoolbox}
\usepackage[font=small,labelfont=bf]{caption} 
\usepackage{hyperref}
\usepackage{sidecap}

\usepackage{booktabs}
\usepackage{pifont}
\usepackage{graphicx}
\usepackage{subfig}
\usepackage{xspace}
\usepackage{wrapfig}
\usepackage{float}
\definecolor{mygreen}{RGB}{15, 157, 88}
\definecolor{myred}{RGB}{219, 68, 55}

\usepackage{hyperref}
\let\svthefootnote\thefootnote
\newcommand\freefootnote[1]{%
  \let\thefootnote\relax%
  \footnotetext{#1}%
  \let\thefootnote\svthefootnote%
}

\newcommand{\ours}{ArrayBot\xspace}

\title{\LARGE \bf
\ours: Reinforcement Learning for Generalizable Distributed Manipulation through Touch
}
\author{
Zhengrong Xue$^{*\,1,2,3}$,
Han Zhang$^{*\,1,2}$,
Jingwen Cheng$^{1}$,
Zhengmao He$^{2}$,\\
Yuanchen Ju$^{2}$,
Changyi Lin$^{2}$,
Gu Zhang$^{2,4}$,
Huazhe Xu$^{\dagger \,1,2,3}$
}

\begin{document}


\twocolumn[{%
\renewcommand\twocolumn[1][]{#1}%
\maketitle
\vspace{-5mm}
\begin{center}
\url{https://steven-xzr.github.io/ArrayBot}
\vspace{3mm}
\end{center}
}]



\thispagestyle{empty}
\pagestyle{empty}

\freefootnote{${}^*\,$Equal Contribution. ${}^\dagger \,$Corresponding Author.}
\freefootnote{$^1\,$Tsinghua Embodied AI Lab, IIIS, Tsinghua University. $^2\,$Shanghai Qi Zhi Institute. $^3$\,Shanghai AI Lab. $^4$\,Shanghai Jiao Tong University.}
\freefootnote{Contact:\tt\small{ xzr23@mails.tsinghua.edu.cn, huazhe\_xu@mail.tsinghua.edu.cn}.}

\begin{abstract}
    We present \ours, a distributed manipulation system consisting of a $\mathbf{16\,\times\,16}$ array of vertically sliding pillars integrated with tactile sensors. Functionally, \ours is designed to simultaneously support, perceive, and manipulate the tabletop objects. Towards generalizable distributed manipulation, we leverage reinforcement learning (RL) algorithms for the automatic discovery of control policies. In the face of the massively redundant actions, we propose to reshape the action space by considering the spatially local action patch and the low-frequency actions in the frequency domain. With this reshaped action space, we train RL agents that can relocate diverse objects through tactile observations only. 
    Intriguingly, we find that the discovered policy can not only generalize to unseen object shapes in the simulator but also have the ability to transfer to the physical robot without any sim-to-real fine-tuning. Leveraging the deployed policy, we derive more real-world manipulation skills on \ours to further illustrate the distinctive merits of our proposed system.
\end{abstract}



\section{Introduction}

The notion of robotic manipulation~\cite{billard2019trends,kroemer2021review} easily invokes the image of a biomimetic robot arm or hand trying to grasp tabletop objects and then rearrange them into desired configurations inferred by exteroceptive sensors such as RGBD cameras. To facilitate this manipulation pipeline, the research community has made tremendous efforts in either how to determine steadier grasping poses in demanding scenarios~\cite{kerrevo,zhou2023learning,zhu2022sample,shao2020unigrasp,james2020slip} or how to understand the exteroceptive inputs in a more robust and generalizable way~\cite{shi2021skeleton,yuan2022sornet,simeonov2022neural,xue2022useek,radosavovic2023real,wen2022you}. Acknowledging these progresses, we attempt to bypass the above challenges by advocating ArrayBot --- a reinforcement learning (RL)  ~\cite{lillicrap2015continuous,schulman2017proximal} driven and tactile observation only~\cite{khandate2022feasibility,guzey2023dexterity,yin2023rotating} \textit{distributed manipulation}~\cite{bohringer2000distributed} system, where the objects are manipulated via numerous contact points.

Conceptually, the hardware of \ours is a $16 \times 16$ array of vertically sliding pillars, each of which can be independently actuated, leading to a $16 \times 16$ action space.
Functionally, the pillars beneath a tabletop object can support its weight and at the same time cooperate to lift, tilt, or even translate it through proper motion policies. To equip \ours with proprioceptive sensing, we integrate each pillar with a slim and low-cost Force Sensing Resistor (FSR) sensor, allowing the robot to ``feel'' the object when lack of external visual inputs. Thanks to its distributed nature, \ours is flexible in size, inherently supports manipulation in parallel, and has the potential to manipulate objects times larger than the size of its end-effector.

\begin{figure}[t]
    \centering
    \includegraphics[width=\linewidth]{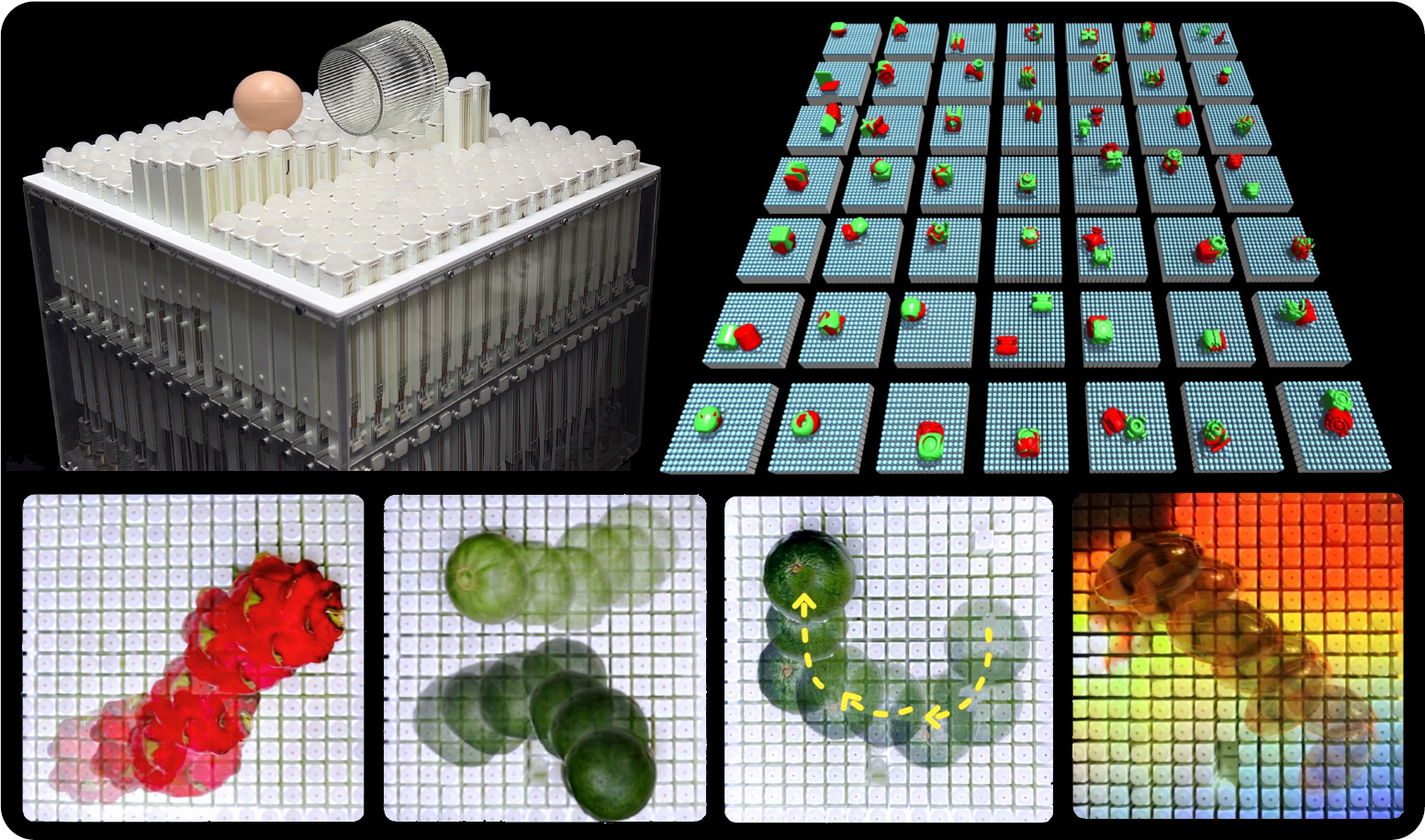}
    \vspace{-4mm}
    \caption{We present \ours, a distributed manipulation system. With the aim of generalizable manipulation, we train RL agents on the simulated \ours where the only accessible observation is the tactile information. Afterwards, we deploy the learned control policy to the physical robot, and showcase the bird's-eye view of the trajectories for real-world manipulation tasks: relocating novel-shaped objects, manipulating two objects in parallel, trajectory following, and manipulation under visual degradations. Please refer to the videos on our \href{https://steven-xzr.github.io/ArrayBot/}{project website}.}
    \vspace{-3mm}
    \label{fig:teaser}
\end{figure}

Previous works for distributed manipulation show up in the names of actuator array~\cite{bohringer1995sensorless,luntz2001distributed,thompson2021towards,patil2022linear}, smart surface~\cite{barr2013smart,dang2016electromagnetic}, or auxiliary functions of tangible user interface~\cite{pangaro2002actuated,follmer2013inform,leithinger2015shape}. Despite their promises to manipulate tabletop objects, they heavily depend on pre-defined motion primitives to fit the specific designs of the systems.
With the configurations (e.g., shapes, positions, etc.) of the manipulated objects altering, human-determined rules may require fine-tuned parameters or even a thorough redesign.
Towards distributed manipulation enjoying better generalizability and versatility, we explore the feasibility of applying model-free RL~\cite{lillicrap2015continuous,haarnoja2018soft,schulman2017proximal} to the automatic discovery of control policies. However, compared with popular manipulators such as arms or hands, controlling \ours in its 2D-array action space can be extremely challenging because the massive redundancy of the actions makes the trial-and-error process hopelessly inefficient. 

\begin{figure*}[t]
    \centering
    \includegraphics[width=0.88\linewidth]{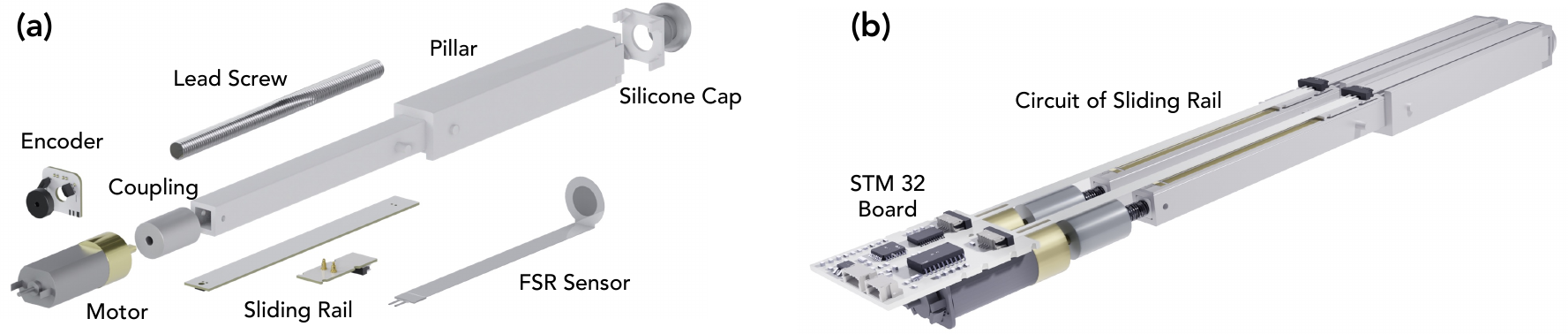}
    \caption{The hardware of \ours is a $16 \times 16$ array of vertically sliding pillars. (a) The exploded view of an atom unit, which consists of the actuator, the pillar, and the end-effector. (b) Every two atom units are assembled with one STM32  board as a modular unit.}
    \vspace{-3mm}
    \label{fig:hardware}
\end{figure*}

In awareness of its redundancy, we propose to reshape the action space with the objective to strengthen its inductive bias towards more favorable actions for distributed manipulation.
To start with, we explicitly restrict the extend of the valid action space to the $5 \times 5$ \textit{Local Action Patch} centered around the object.
Meanwhile, we propose the idea of considering \textit{Actions in the Frequency Domain} via 2D Discrete Cosine Transform (DCT)~\cite{ahmed1974discrete}.
Our intuition is that each channel in the frequency domain processes a spatially global horizon, so a frequency-domain perspective may help promote the collaborations among spatially neighboring pillars.
On top of the frequency transform, we further perform \textit{High Frequency Truncation} on the action channels. The rationale behind is that lower-frequency actions may correspond to actions with emergent semantics, e.g., the DC channel implies lifting, and the base frequency implies tilting.

With the reshaped action space ready, we set up the simulated \ours in the Isaac Gym simulator~\cite{makoviychuk2021isaac} and train model-free RL agents~\cite{schulman2017proximal} that can respectively lift and flip a cube. Going beyond the reach of simple non-generalizable motion skills, we manage to acquire one generalizable policy discovered by RL that is agnostic of both object shapes and visual observations, but could transport diverse-shaped previously unseen objects from and to arbitrary positions via touch sensing alone. Interestingly, we find it might be easier than expected to deploy the policy trained on the simulated \ours to the real-world machine. Without any sim-to-real fine-tuning, the averaged success rate of the \textit{general relocate-via-touch} policy is tested to reach $74\%$ on a batch of unseen objects, leading to an already decent baseline.

Leveraging the \textit{general relocate-via-touch} policy deployed to the real world, we further illustrate the characteristic merits of \ours by presenting the following derived manipulation skills --- trajectory following by iteratively calling the relocation policy, manipulating multiple objects in parallel thanks to its distributed nature, and manipulation under visual degradations due to no visual observations at all. As the ending of this paper, we also envision the potential applications that \ours may empower in the future in both industrial and household scenarios.

\section{A Sketch for the Hardware Design}

The hardware of \ours can be perceived as a $16 \times 16$ array of vertically sliding pillars. Each atom unit from down to up consists of an actuator, a rectangular pillar whose length-width-height is $16 \times 16 \times 200$ mm, a slim and low-cost Force Sensing Resistor (FSR) sensor that measures the pressure, and a silicone hemispheric end-effector that protects the tactile sensor and increases the frictions. 


%



\noindent\textbf{Actuator.}
The left side of Figure~\ref{fig:hardware}(a) is the actuator, which is a DC gear motor. The rotational motion of the motor are converted into the translational motion of the pillar through a screw structure. A magnetic encoder is installed on the rotating shaft to calculate the angle and angular velocity of the motor, which are ultimately mapped into the vertical position and speed of the joints. The effective range of each vertically prismatic joint is $55$ mm, whose maximum motion speed is $53$ mm/s. The movements of the joints are controlled by STM32 microcontrollers, and the target actions are executed via positional PID control.


\noindent\textbf{End-effector.}
The right side of Figure~\ref{fig:hardware}(a) is the end-effector, which has a silicone semi-spheric cap and an FSR tactile sensor whose effective measuring range is $10{\footnotesize \sim}200$ grams. 
When an object is placed on the end-effector, its pressure is transmitted through the silicone cap to the FSR sensor. 
Since the pedestal which the sensor rests upon is consistently sliding, we would better avoid the use of wires when installing the sensor. Thus, we design a conductive sliding rail embedded underneath the sensor connector for both power supply and signal transmission.

\noindent\textbf{Assembled modular unit.}
As shown in Figure~\ref{fig:hardware}(b), every two atom units are assembled into a modular unit so as to facilitate assembly and make the most of the STM32 microcontrollers.
To diminish signal inferences, we employ $4$ independent CAN buses for the communication between the microcontrollers and the desktop host, where each CAN bus takes charge of $32$ modular units (i.e., $64$ pillars). During operation, each STM32 board receives and processes the CAN commands from the desktop, and sends two PWM signals respectively to the two motors under its control.

\section{Action Space Reshaping}

The central challenge against the employment of RL for distributed manipulation comes from the massive redundancy in its unconventional action space. 
In this section, we present a series of techniques to reshape the action space of \ours so that it is more favored for distributed manipulation.

\noindent\textbf{Local Action Patch.}
The action space of \ours is in the shape of a $16 \times 16$ array. Given the fact that the actuators far away from the object could not make any physical impact, we only consider a $5 \times 5$ Local Action Patch~(LAP) centered around the object.
So far, an untouched detail is how to determine the center of the local patch. If the ground-truth object positions are accessible in the simulator, we simply select the actuator that is closest to the center of the object as the center of the LAP.
Otherwise, we estimate the object position through touch. More specifically, the noisy readings of the $16 \times 16$ FSR sensor array are binarized to enhance robustness, giving a tactile map that indicates contact conditions.
Since the measurements between $10{\footnotesize \sim}13$ grams are found to be unreliable sometimes, we set the rule that if the reading of any FSR sensor is larger than 13 grams, it is regarded to be in contact with the object above it.
The geometric center of all contact points is calculated as the estimated position of the manipulated object.

\noindent\textbf{Actions in the Frequency Domain.}
As clearly figuring out the impact of every individual contact is virtually insolvable~\cite{ajay2018augmenting,fazeli2020fundamental} in a contact-rich environment, we focus on the collective impacts of many actuators instead~\cite{follmer2013inform,thompson2021towards,patil2022linear}.
On a methodology level, we propose to learn Actions in the Frequency Domain since each frequency-domain component may have a global impact in the spatial domain. 
Hence, rather than directly predict a flattened $25$-dim delta positions, the policy network outputs a $25$-dim delta frequencies. Subsequently, the $25$-dim output is unflattened to the shape of $5 \times 5$ and then post-processed by a 2D inverse Discrete Cosine Transform (iDCT)~\cite{ahmed1974discrete} operator to produce the $5 \times 5$ action in the spatial domain.

\begin{wrapfigure}[14]{r}{0.22\textwidth}%
    \centering
    \vspace{-13pt}
    \includegraphics[width=0.22\textwidth]{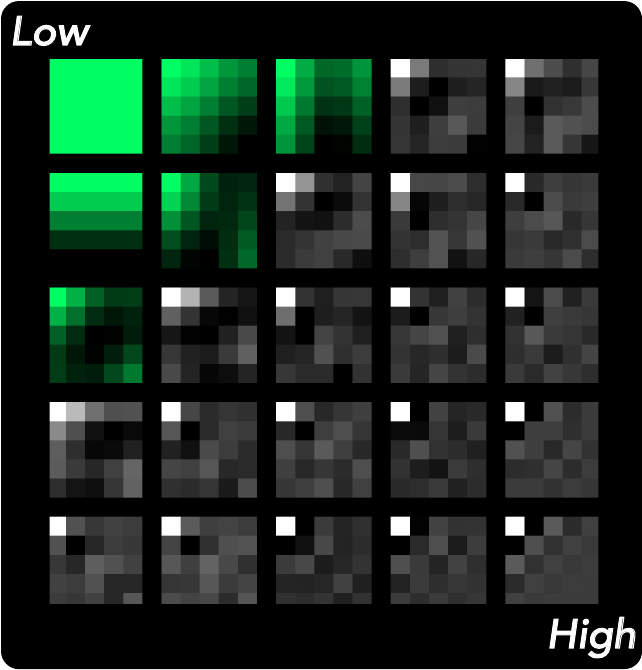}
    \vspace{-13pt}
    \caption{The visualization of a $5 \times 5$ 2D DCT map. We select the lowest $6$ frequency channels marked in green.}
\end{wrapfigure}

\noindent\textbf{High Frequency Truncation.}
Besides a latent inductive bias towards collaborations, the frequency domain also provides a valuable point of view to re-inspect the redundancy of the actions. Intuitively, lower-frequency channels lead to smooth planar surfaces with semantics that are likely to correspond to emergent motion primitives. For instance, the DC channel implies lifting, and the base frequency implies tilting.
In comparison, high-frequency channels mainly represent fine texture information, whose impact on manipulation is relatively limited. 
Based on these observations and inspired by image compression methods such as JPEG~\cite{wallace1991jpeg}, we propose to truncate the high-frequency channels of the actions. Ultimately, the policy network is designed to output a $6$-dim prediction, which is used to fill the lowest $6$ frequency channels of the entire predicted action in the frequency domain. To acquire a full $25$-dim action ready for the inverse frequency transform, we simply zero-pad the rest $19$ channels of higher frequencies. 




\begin{figure*}[!t]
    \centering
    \includegraphics[width=0.95\linewidth]{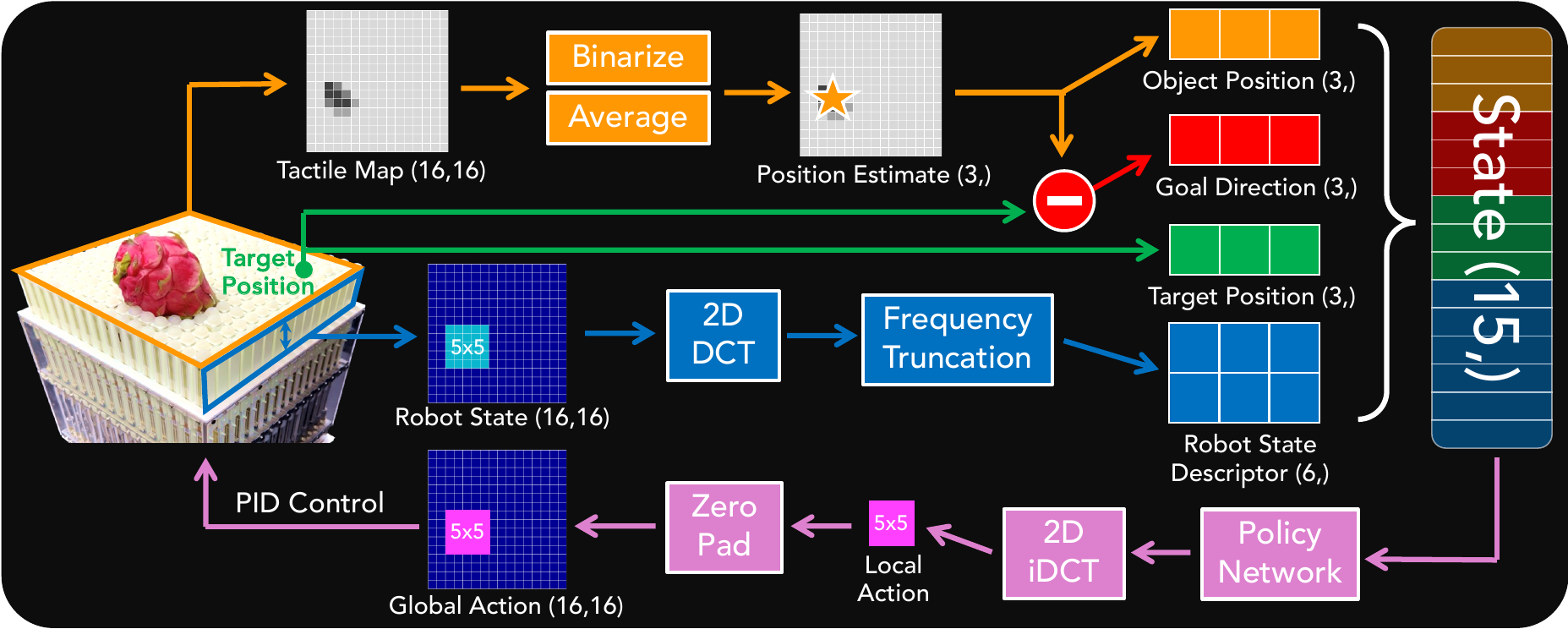}
    \caption{An overview of the RL framework on \ours for \textit{general relocate-via-touch}. The state is the combination of the estimated object position, the specified target position, the residual goal direction, and the robot state in the frequency domain. Exempt from any visual inputs, the states are inferred from purely proprioceptive observations of the robot joint configuration and the tactile sensor array.}
    \label{fig:pipeline}
    \vspace{-0.6em}
\end{figure*}

\section{Learning the Control Policies}

\subsection{Simulator Setup}
The physical simulation of contact-rich interactions could be time-consuming. To produce sufficient samples in an efficient way so as to feed data-hungry RL algorithms, we build the simulated environment in the Isaac Gym~\cite{makoviychuk2021isaac} simulator. The frequency of the physical simulation steps is $50$ Hz. Due to the mechanical speed limit, we set the frequency of RL control to be $5$ Hz.
Since the RL algorithm considers the binarized outcome of the tactile sensor, we simply retrieve the information from the contact buffer of the simulator as the simulation of tactile sensors.

\vspace{-0.2em}

\subsection{Environments}
To verify the effectiveness of the proposed action space, we devise the environment of \textit{lifting} where \ours is asked to raise up a cubic block, and \textit{flipping} where \ours is asked to flip the same block by $90$ degrees. To explore the full potential of our system, we also study a more challenging setting of \textit{general relocate-via-touch} where \ours is asked to relocate unseen-shaped objects from and to any arbitrary positions through tactile sensing only.


\noindent\textbf{States.} The tasks of \textit{lifting} and \textit{flipping} directly make use of the privileged position and orientation information provided by the simulator. In \textit{general relocate-via-touch}, we consider a more realistic scenario where the only observation is the binarized tactile information. The estimated states for \textit{general relocate-via-touch} are visualized in Figure~\ref{fig:pipeline}.


\noindent\textbf{Rewards.} The tasks of \textit{lifting} and \textit{flipping} simply consider a dense reward of object height and orientation respectively. In \textit{general relocate-via-touch}, apart from the dense reward of object position, we add one more sparse bonus reward when reaching the goal that would encourage the robot to timely stop the object at the goal position.


\noindent\textbf{Actions.} At each step, the policy outputs a $6$-dim action in the frequency domain, which is post-processed to produce the relative joint configuration on the $5 \times 5$ LAP. 


\noindent\textbf{Resets.} We reset the episode if the tabletop object moves out of the border or the episode length reaches $100$ steps. With the existence of the $5 \times 5$ LAP, we request that the center of the object should locate on the central $11 \times 11$ patch. 

\noindent\textbf{Manipulated objects.}
In \textit{lifting} and \textit{flipping}, we manipulate an $8 \times 8 \times 8$ cm cubic block which can be roughly supported by a $4 \times 4$ array of actuators.
In \textit{general relocate-via-touch}, we train an RL agent that is generalizable to shape variance by sampling $128$ different shapes from the EGAD~\cite{morrison2020egad} training set and then re-scaling them. At test time, we evaluate the performance of the generalizable agent on the EGAD test set with a total of $49$ unseen object shapes.
For fast and accurate collision detection in the simulator, we perform V-HACD~\cite{mamou2016volumetric} convex decomposition to all of the object shapes before loading them into the simulator.


\subsection{Training the RL Agents}

For all the tasks, we train proximal policy optimization (PPO)~\cite{schulman2017proximal} agents on $128$ parallel Isaac Gym environments for the automatic discovery of control policies. 
Notably, all of the $128$ parallel environments for \textit{general relocate-via-touch} involve mutually different object shapes. With a state space agnostic of the object shapes at all, the agent receives mixed types of dynamics. This forces the agent to discover a policy that is as universal as possible for all the shapes in the dataset, which is likely to enhance the agent's generalizability towards unseen object shapes.

\begin{figure*}[t]
    \centering
    \includegraphics[width=0.98\linewidth]{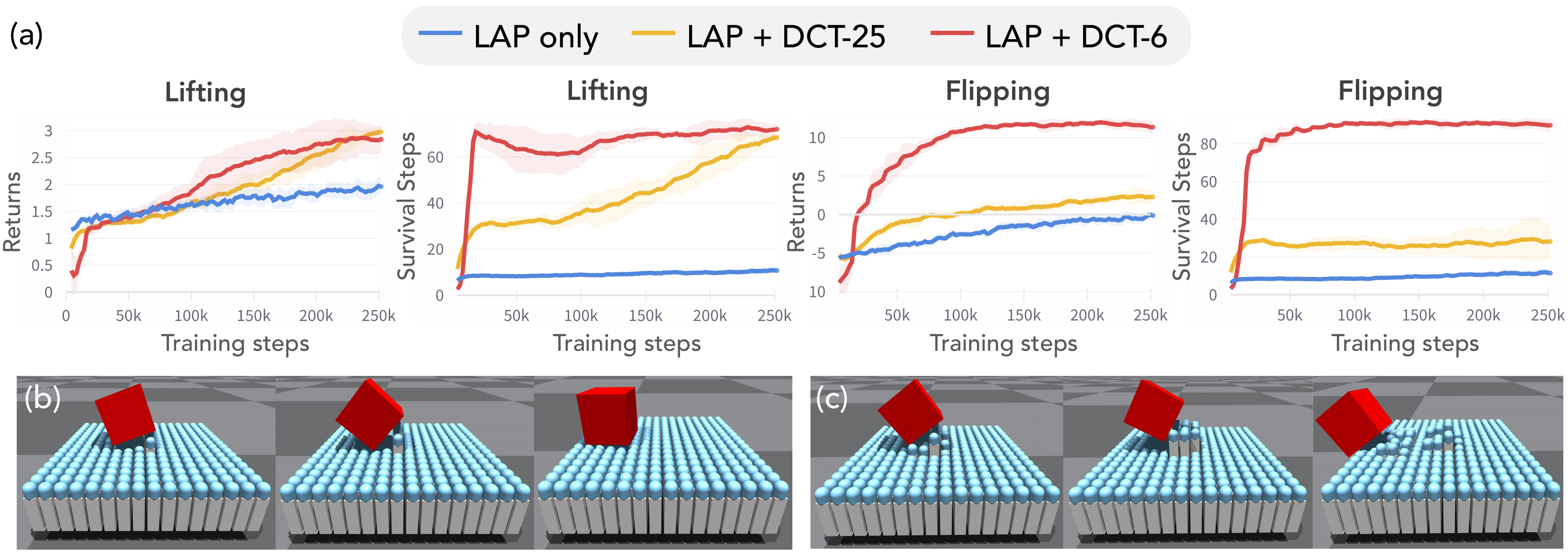}
    \caption{(a) The training curves in terms of episode returns and survival steps. The results are averaged on $5$ seeds. The shaded area stands for the standard deviation. (b)(c) The example trajectories of the policies learned by (b)~\textit{LAP+DCT-6} and (c)~\textit{LAP only} for \textit{flipping}.}
    \vspace{-2mm}
    \label{fig:lift-tilt}
\end{figure*}

\subsection{Simulated Experiments for Lifting and Flipping}

\noindent\textbf{Metrics.}
In \textit{lifting} and \textit{flipping}, we compare the averaged accumulated returns and the survival steps of each episode. The survival step refers to the steps an object could stay on the robot without falling, whose maximum is $100$. 

\noindent\textbf{Compared methods.}
To study the necessity of our reshaped action space, we train the same PPO algorithm in the following action spaces: (i) \textit{LAP only} in the spatial domain; (ii) \textit{LAP+DCT-25} that preserves all of the $25$ channels in the frequency domain; and (iii) \textit{LAP+DCT-6} that considers only the $6$ lowest-frequency channels.

\noindent\textbf{Results.}
The learning curves of both tasks are shown in Figure~\ref{fig:lift-tilt}(a). In both tasks, the DCT-based approaches have a significant advantage over the one trained in the spatial action space in terms of both total returns and survival steps. Further, the \textit{DCT-6} method survives longer and behaves better than \textit{DCT-25}, especially in the more challenging task of \textit{flipping}, echoing the intuition that low-frequency patterns lead to more steady actions.
By visualizing the \textit{flipping} trajectories in Figure~\ref{fig:lift-tilt}(b)(c), we find that \textit{LAP only} hacks the environment and learns to gain rewards by throwing the block off the robot in a rolling way. In comparison, the actions of \textit{LAP+DCT-6} are more gentle and reasonable, which explains its better performance and longer survival time.


\subsection{Simulated Experiments for General Relocate-via-Touch}
\noindent\textbf{Metrics.} We report the success rate of relocation on the EGAD~\cite{morrison2020egad} test set containing $49$ unseen objects. An episode is judged to succeed if the object reaches the target position and insists for at least $1$ second. The results are averaged over $200$ trials with random initial and target positions.

    

\begin{figure}
    \centering
    \includegraphics[width=0.49\textwidth]{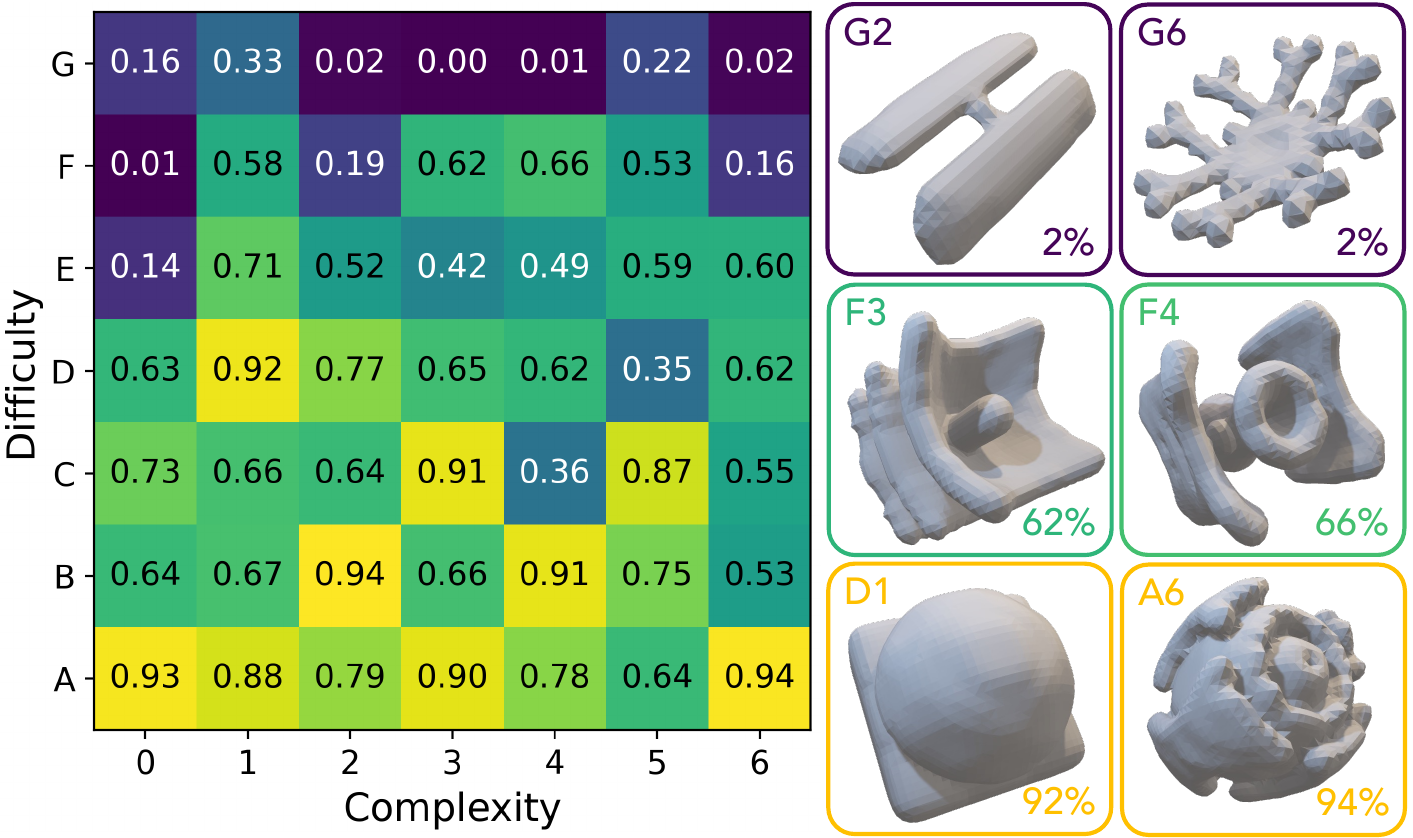}
    \caption{(Left) The success rates of the \textit{general relocate-via-touch} policy evaluated on the previously unseen EGAD~\cite{morrison2020egad} testset in the simulated environment. (Right) Visualization of some representative objects in the dataset and their corresponding success rates.}
    \label{fig:relocation-success-rate}
    \vspace{-3mm}
\end{figure}

\noindent\textbf{Results.}
The form of success rates shown in Figure~\ref{fig:relocation-success-rate} follows the same taxonomy as the EGAD dataset where the grasping difficulty and shape complexity of objects is alphabetically and numerically sorted.
Our findings are as follows:
(i) \ours achieves relatively high success rates on the majority of easy (Level A${\footnotesize \sim}$D) objects even when the shape complexity is high (e.g., A6). This is probably because the shapes of these objects can be largely enveloped by some simple convex shape primitives such as sphere or cube, which can be more easily handled by the mixture of sliding and rolling policy that RL discovers.
(ii) \ours is generally not good at manipulating objects that are extremely hard for grasping (i.e., Level G). In contrast to easier instances in the dataset, the shape outlines of objects on Level G are typically very flat and concave (e.g., G2 \& G6), which are unfriendly to both grippers and \ours. 
(iii) Specific objects hard to grasp are relatively easy for \ours (e.g., F3 \& F4), which implies \ours might have complementary ability compared with existing manipulators such as arms or hands.

Considering all the shape variations are handled by the same policy and none of the objects is shown to the RL agent at training time, we believe \ours has demonstrated much capability of generalizable manipulation, and thus is ready for the policy deployment on the physical machine.



\section{Deploying the Control Policies}
\subsection{Zero-Shot Sim-to-Real Transfer}
Intriguingly, we find that the \textit{general relocate-via-touch} policy trained on the EGAD dataset in the simulator can be directly deployed to the physical robot without further sim-to-real fine-tuning. To strengthen the significance of our finding, we exempt the usage of domain randomization~\cite{tobin2017domain}, which is accepted as a critical technique for the successful sim-to-real transfer of many RL-discovered policies~\cite{peng2018sim,andrychowicz2020learning}.

Intuitively, there are two main sources that the sim-to-real gaps originate from: perception and motion dynamics. Our selection of proprioceptive observations and binarized tactile measurements keeps the discrepancies in perception at a low level. Meanwhile, both the diverse shapes in the training stage and the massive redundancy in the action space contribute to the resilience towards the shift in dynamics.

\subsection{Experiments on the Physical Robot}
For quantitative evaluation of the \textit{general relocate-via-touch} policy deployed to the physical robot, we casually place the manipulated object in an initial position and require the policy to translate it to a specified target position. We select five diverse-shaped daily-life objects for evaluation: melon, pineapple, dragon fruit, rugby, and cube. For each object, we run $10$ trials and then report the success rate. 

For ablation and comparison, we design a hard-coded policy where the pillars are scripted into the shape of a ``cage'' carrying the manipulated object. We manually set the moving trajectory of the ``cage'' and hope that the object can be manipulated alongside the pre-programmed course.

The success rates are reported in Table~\ref{tab:success-rate}.
Overall, the performance of the RL-discovered policy is more consistent in the face of object shape variations, and on average outperforms that of the hard-coded policy by $20\%$. We find that the performance of the hard-coded policy is more sensitive to the sizes and shapes of the manipulated objects. If the object happens to fit the space of the ``cage'' (e.g., the rugby), then it can be well manipulated by the hard-coded policy. Otherwise, the object may slide out of the ``cage'', leading to failed trials. Besides, for both hard-coded and RL-discovered policy, we observe a failure mode where some protruding parts of the objects may get stuck into the gaps between the pillars, thus hindering subsequent manipulation.
We expect the problem can be alleviated by hardware iterations equipped with end-effectors whose sizes are reduced so that enhanced manipulation granularity can be provided.

\begin{table}
\centering
\vspace{1.5mm}
\begin{tabular}{lc|cc}
    \toprule
     & Weight (g) & Hard-Coded & RL-Discovered \\
    \midrule
    Melon & 449 & $4/10$ & $9/10$ \\
    Pineapple & 550 & $6/10$ & $8/10$ \\
    Dragon Fruit & 369 & $7/10$ & $8/10$ \\
    Rugby & 244 & $10/10$ & $7/10$ \\
    Cube & 391 & $0/10$ & $5/10$ \\
    \midrule
    Overall & - & $54\%$ & $\mathbf{74\%}$ \\
    \bottomrule
\end{tabular}
\caption{The success rates of the hard-coded and RL-discovered \textit{general relocate-via-touch} policy evaluated on the physical robot. The selected objects at test time are unseen in the training procedure, whose weights and sizes (half-extents) range from $224{\footnotesize \sim}550$\,g and $3{\footnotesize \sim}5.5$\,cm, respectively. }
\vspace{-3mm}
\label{tab:success-rate}
\end{table}


\begin{figure*}[t]
    \centering
    \includegraphics[width=\linewidth]{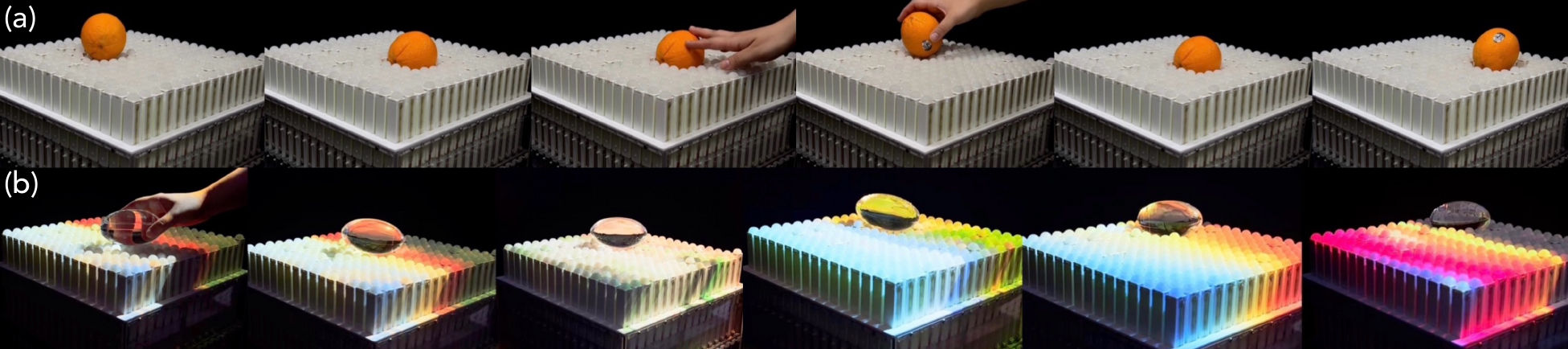}
    \caption{Real-world trajectories of the manipulation tasks showing the robustness of our system to the impacts of (a) unexpected external forces (b) severe visual degradations. For more detailed visual illustrations, please refer to the \href{https://steven-xzr.github.io/ArrayBot/}{project website}.}
    \label{fig:real-world}
\end{figure*}

\subsection{Derived Real-World Manipulation Skills}
Leveraging the \textit{general relocate-via-touch} policy deployed to the real world, we manage to further demonstrate the merits of \ours by presenting the following derived manipulation skills on the physical robot:

\begin{itemize}

\item \textit{Trajectory following.} Trajectory following can be easily achieved by the iterative calls of the \textit{general relocate-via-touch} policy. Note that \ours is more friendly to incremental operations since its efforts are in proportion to the traveled distance. In comparison, slight operations are the same troublesome as longer-range movements for arms or hands since the costs in ``pick'' and ``place'' are constants whatever the scale of the operations.

\item \textit{Manipulating objects in parallel.} Consisting of a large number of actuators, \ours inherently supports parallel manipulation. Thanks to the spatial self-similarity in the mechanical structure, the same control policy automatically adapts to all local patches. Assuming collision-free target trajectories, manipulating objects in parallel is as easy as initializing multiple independent manipulation processes. More complicated collision-involved settings are left for future work.

\item \textit{Manipulation under visual degradations.} We point out the fact that the performance of \ours is unaffected by visual degradations, which is a concrete benefit brought by using tactile-only observations.

\end{itemize}

\section{Related Works}

\noindent\textbf{Distributed manipulation} controls the motion of the target object through numerous points of contact~\cite{bohringer2000distributed}. Composed of an array of stationary unit cells, a distributed manipulation system is able to be scaled in size and inherently support manipulation in parallel. Reviewing the literature, the majority of distributed manipulation systems consist of an array of special-purpose actuators such as vibrating plates~\cite{bohringer1995sensorless}, air jets~\cite{luntz2001air}, roller wheels~\cite{luntz2001distributed}, electromagnets~\cite{pangaro2002actuated} and delta robots~\cite{thompson2021towards,patil2022linear}. While they are designed to be skilled in specific types of manipulation tasks, they are typically not versatile enough and demand elaborately pre-defined motion primitives. Compared with the prototypes in robotics research, the hardware of \ours is more related to the branch of works known as the ``shape-changing tangible user interfaces''~\cite{follmer2013inform,leithinger2015shape,siu2018shapeshift} in the human-computer interface community, where the actuators are vertically prismatic pillars. The simplicity in design not only makes it easier to manufacture; its organized action space also helps open the door to learning motion policies with model-free RL. 
\noindent\textbf{Learning in the frequency domain}
is a concept discussed in a variety of scopes in the machine learning community. In computer vision studies, frequency transformations are adopted to bridge the gap between high-quality images and down-sampled ones~\cite{xu2020learning}, and help achieve more accurate gradient approximations in Binary CNNs~\cite{xu2021learning}.
In the topics of reinforcement learning, the concept of frequency is utilized to boost the efficiency of search-control in model-based architectures~\cite{pan2020frequency}, and represent the characteristic function of returns for distributional RL~\cite{farahmand2019value}. 
In contrast to all the existing works that we have found, frequency transformation is leveraged in \ours for action space reshaping, aiming at reformative sample efficiency in model-free RL.

\section{Discussion on Potential Applications}



In the industrial scenario, \ours could serve as an integrated conveyor and sortation system. Current solutions for automated sortation systems generally rely on a vision-based perception module and an execution module consisting of a robotic arm and a specialized end-effector~\cite{putwall}. Compared with the solutions on the market, \ours could: (i) avoid the challenges in grasping irregular objects by directly sorting objects on the conveyor, (ii) manipulate numerous objects in parallel to boost the sortation efficiency, (iii) operate in the dark or under dramatically varying lighting conditions.

In the household scenario, \ours can facilitate the users by either manipulating the tabletop objects or altering its own shape according to their needs. For instance, when a cellphone placed on \ours receives a new message and vibrates, \ours can detect the vibration through tactile sensors and then push the cellphone to the user's side. This application may also apply to other tools in the daily life.

\section{Limitations and Future Work}

A prominent limitation of the \ours prototype is the non-negligible 16 mm side length of the end-effector (EE) and the 4 mm gap in between. Coarse-grained EEs not only are to blame for the main failure modes in the relocation task, but also makes it impossible to manipulate objects smaller than the EE itself.
Meanwhile, very lightweight objects might not be detected by the FSR tactile sensor due to its limited measuring precision. The problem is deteriorated by the fact that pressures might be shared by multiple sensors. In the future, we will make efforts to reduce the EE size and gap space as well as improve the precision of the tactile sensor so that we can explore the greater potential of \ours for more dexterous manipulation tasks.

\section{Conclusion}

We present \ours, an RL-driven distributed manipulation system via tactile sensing. In the simulator, we develop policies for lifting, flipping, and generalizable relocate-via-touch in the proposed reshaped action space, and then successfully deploy the policies to the physical robot without much sim-to-real fine-tuning. Leveraging the deployed relocation policy, we showcase the distinctive merits of our \ours through derived real-world manipulation skills. 


\section*{Acknowledgement}
We would like to give special thanks to Mr. Yuanhong Wang and his group of colleagues from Kunshan Qiaotuo ModelTech Co., Ltd. for their contributions to the manufacture of the hardware. We also thank Yitong Wang for hardware maintenance, Tianyu Huang and Yao Lu for elegant graphic rendering of the hardware design, and Yuzhe Qin for his advice on the Isaac Gym simulator.
Huazhe Xu is supported by National Key R\&D Program of China (2022ZD0161700).

\bibliographystyle{abbrv}
\bibliography{main.bib}

\newpage
\onecolumn
\appendices
\section{More Details in Hardware Design}
\label{apd:hardware}
\subsection{Robotic Frame Design}
\begin{figure}[htp]
    \centering
    \includegraphics[width=\textwidth]{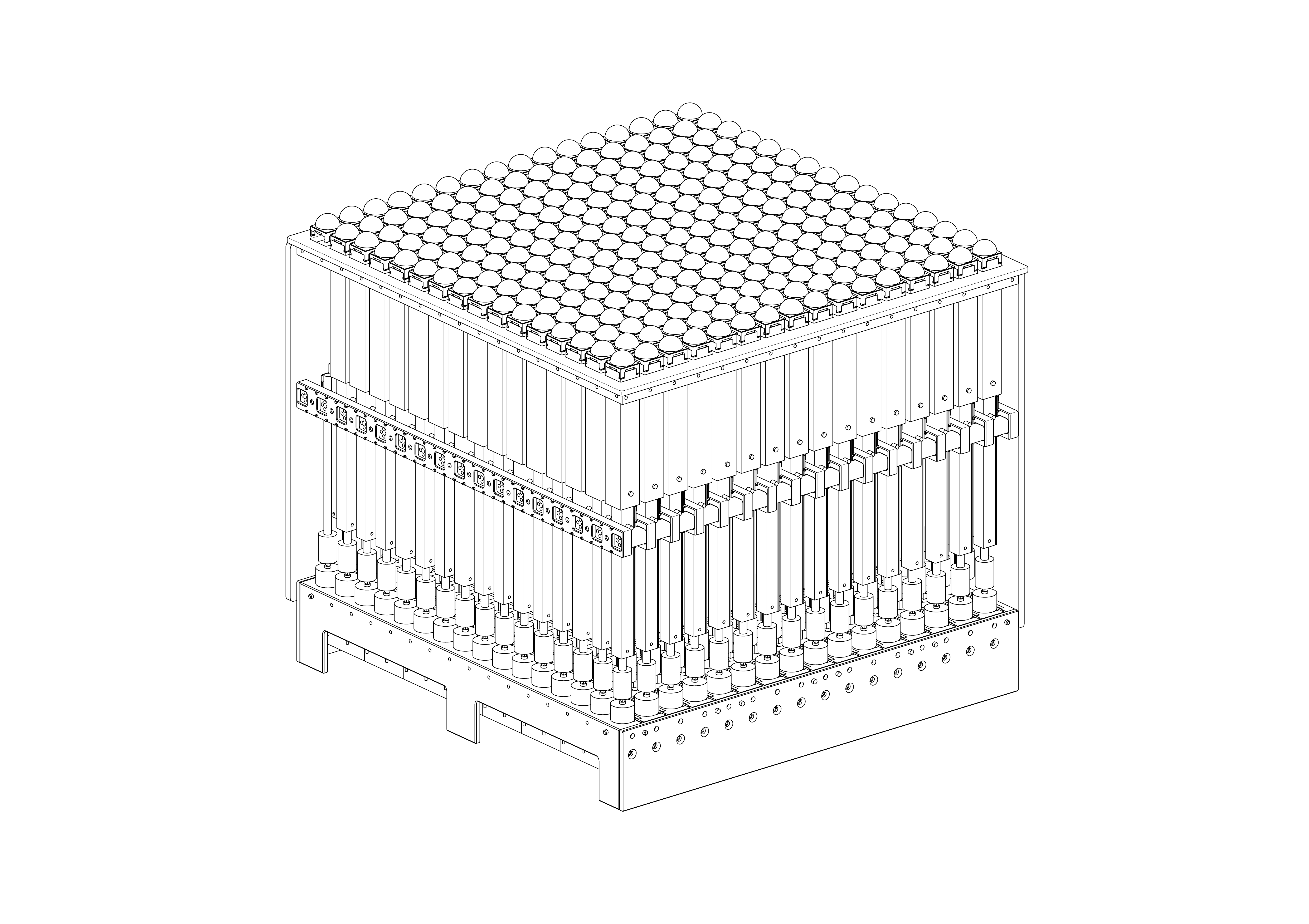}
    \caption{The robotic frame with a grid cover, a transparent shell, and an iron base.}
    \label{fig:16times16_line}
\end{figure}

The robotic frame is shown in Figure ~\ref{fig:16times16_line}. The grid cover on top is made of Acrylonitrile Butadiene Styrene plastic~(ABS), using Computerized Numerical Control~(CNC) for fabrication. The four surrounding sides of the shell are made of acrylic. The motor brackets and column supports are made of aluminum alloy, using CNC for production. The base is made of iron, which is welded and painted.
\subsection{Design and Fabrication of Pillars}
\begin{figure}[htp]
    \centering
    \includegraphics[width=\textwidth]{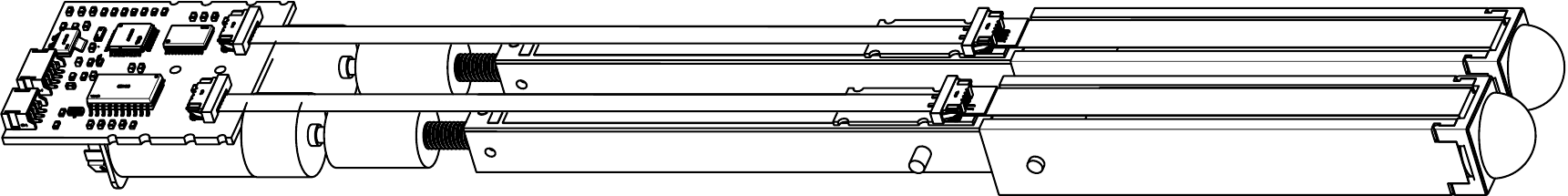}
    \caption{A module of two pillars controlled by a STM32 board.}
    \label{fig:1times2_line}
\end{figure}
\textbf{Silicone cap.} To improve the transmission of force to the sensor, we develop a silicone cap with a 30A hardness. This cap is produced through vacuuming and heat setting processes. The CNC technique is also employed for producing the mold of the silicone cap.

\textbf{Tactile sensor.} We employ the force-sensitive resistor~(FSR) sensor model RXD1016, which has a measuring range of 200 g, a resistance range of 0.5k-10k ohms, and an effective detection area of a circular shape with a diameter of 10 mm. The FSR sensor measures the normal force, and we convert the sensor resistance to voltage through a voltage divider in series. We utilize the microcontroller's Analog-to-Digital Converter~(ADC) port to read the values, with the perceived pressure being directly proportional to the voltage. As the STM32F042 board's ADC can discern up to 0.8 mV voltage, the theoretical trigger value of the sensor is 10.36 g.
In order to mitigate the impact of components such as the silicone cap on assessing the contact condition, we conduct a calibration prior to usage. This process uses the no-load voltage as the new reference zero point to evaluate the contact status.

\textbf{Actuator.} The DC gear motor is CHR-GM16-050ABHL, with a rated voltage of 12 V, a maximum power of 3.3 W, and a no-load speed of 800 rpm. The screw has a pitch of 1 mm and a lead of 4 mm. When the motor completes one rotation, a single column can move up by 4 mm.


\textbf{Microcontroller.} We select the STM32F042 microcontroller due to its inclusion of an ADC interface, a CAN interface, and an ample number of pins. Given its acceptable level of performance and relatively affordable price, this microcontroller is satisfactory for controlling \ours.

\textbf{Power. }The robot is powered by an adjustable DC voltage source, and the power supply outputs 12 V and 5 V respectively to control the motor and the microcontroller.

\section{More details in RL training}
\label{apd:RL-training}

\subsection{RL Algorithm}
Our policy and value functions are separate neural networks with hidden layers of sizes [256, 256, 128]. All the inputs are normalized between 0 to 1 before feeding to the networks. We use ReLU as the activation function. We also list the hyper-parameters of PPO~\cite{schulman2017proximal} in Table \ref{tab:hyperparameters}. The states and rewards are summarized in Table ~\ref{tab:states} and Table ~\ref{tab:rewards}.

\subsection{Training}
Throughout the training phase, reinforcement learning is conducted on GPU, and the simulation is performed on CPU. Acceleration of the simulation is facilitated by using a multi-core parallel method across 128 environments. The frequency for simulation is 50 Hz, while the control frequency is 5 Hz. Training on an A40 graphics card, convergence is achieved after 10 minutes.

\begin{table}[htp]
    \begin{center}
        \caption{A summary of all the states used in all the tasks.}
        \label{tab:states}
        \begin{tabular}{ccc}
            \toprule
            Parameter & Definition & Dimension \\
            \midrule
            
            \multicolumn{3}{c}{\framebox[13cm]{lifting}}\\
            \textbf{$p$} & object position in World Frame & 3 \\

            \multicolumn{3}{c}{\framebox[13cm]{flipping}}\\
            \textbf{$\theta$} & object orientation with axis-angle representation in world frame & 3 \\
            
            \multicolumn{3}{c}{\framebox[13cm]{general relocation}}\\
            
            \textbf{$p_{t}$} & object position in world frame & 3 \\
            \textbf{$p_{g}$} & object target position in world frame & 3 \\
            \textbf{$p_{d}$} & position difference & 3 \\
            \textbf{$q$} & robot joint configuration in the frequency domain & 6 \\
            
            \bottomrule
        \end{tabular}
    \end{center}
\end{table}

\begin{table}[htp]
    \begin{center}
        \caption{A summary of reward functions used this various tasks.}
        \label{tab:rewards}
        \begin{tabular}{ccc}
            \toprule
            Expression & Definition & Coefficient \\
            \midrule
            
            \multicolumn{3}{c}{\framebox[13cm]{lifting}}\\
            \textbf{$h_t - h_{t-1}$} & object height difference & 100 \\

            \multicolumn{3}{c}{\framebox[13cm]{flipping}}\\
            \textbf{$ - \left \| \theta_t \cdot \theta_g^{-1} \right \| + \left \| \theta_{t-1} \cdot \theta_g^{-1} \right \|$} & object-to-goal orientation difference & 10 \\
            
            \multicolumn{3}{c}{\framebox[13cm]{general relocation}}\\
            
            \textbf{$ - \left \| p_t - p_g \right \| + \left \| p_{t-1} - p_g \right \|$} & object position difference & 100 \\
            \textbf{$\mathbb{1} \{ \left \| p_t - p_g \right \| < \epsilon \}, \epsilon=2\mathrm{cm}$} & Goal Reaching Indicator & 1 \\
            
            \bottomrule
        \end{tabular}
    \end{center}
\end{table}

            
            


\begin{table}
    \begin{center}
        \caption{Training Hyper-parameters}
        \label{tab:hyperparameters}
        \begin{tabular}{ccc}
            \toprule
            Hyperparameter & Value\\
            \midrule
            Discount factor & 0.99\\
            GAE parameter & 0.95 \\
            Learning rate & 3e-4 \\
            Environments & 128 \\
            Optimizer & Adam\\
            Normalize input & True\\
            Normalize value & True\\
            Normalize advantage & True\\
            \bottomrule
        \end{tabular}
    \end{center}
\end{table}

\section{More Details in Real-world tasks}
\label{apd:tasks}
\begin{figure}[htp]
    \centering
    \includegraphics[width=\textwidth]{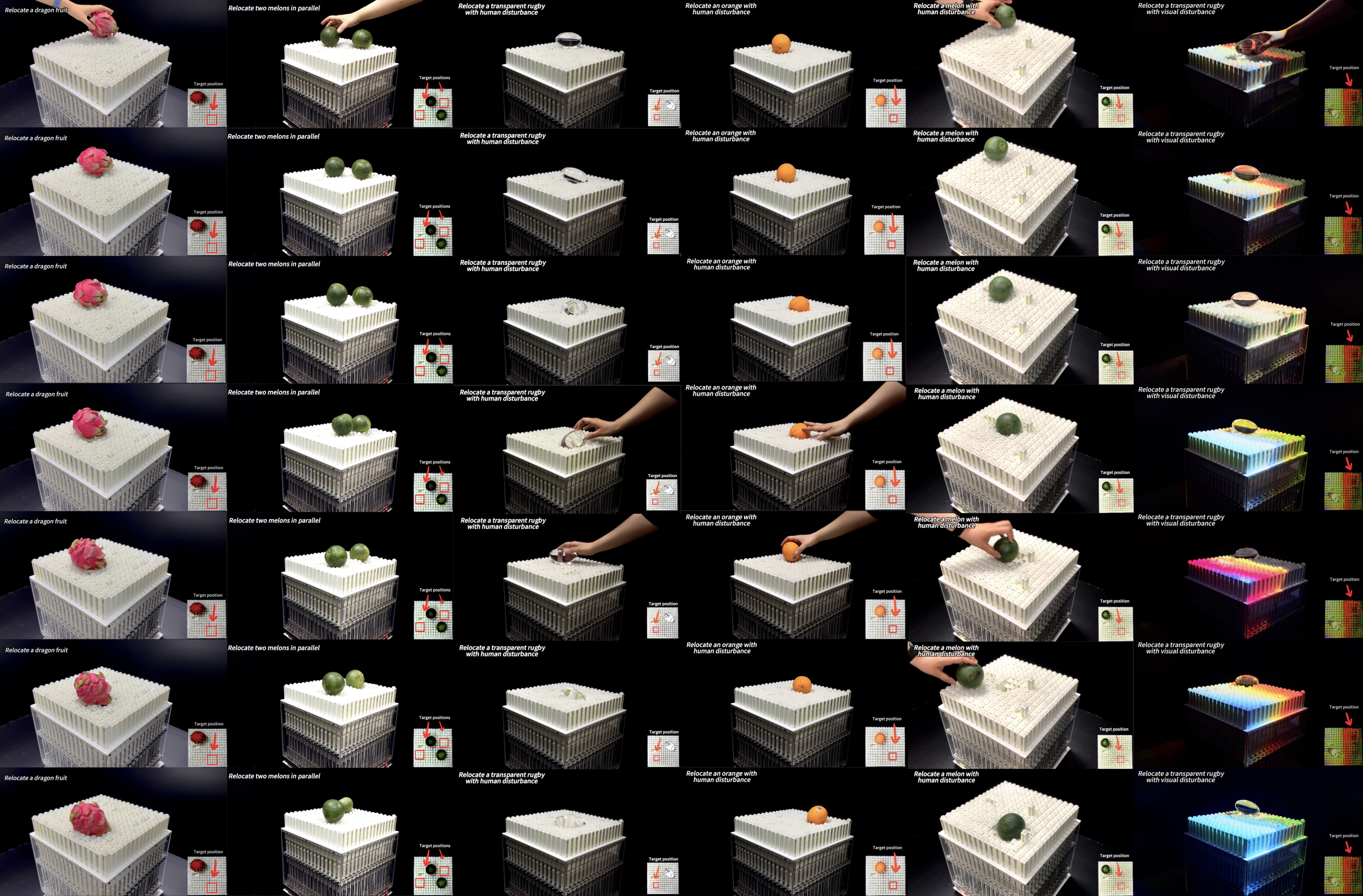}
    \caption{Applications empowered by RL}
    \label{fig:app-RL}
\end{figure}

\begin{figure}[htp]
    \centering
    \includegraphics[width=\textwidth]{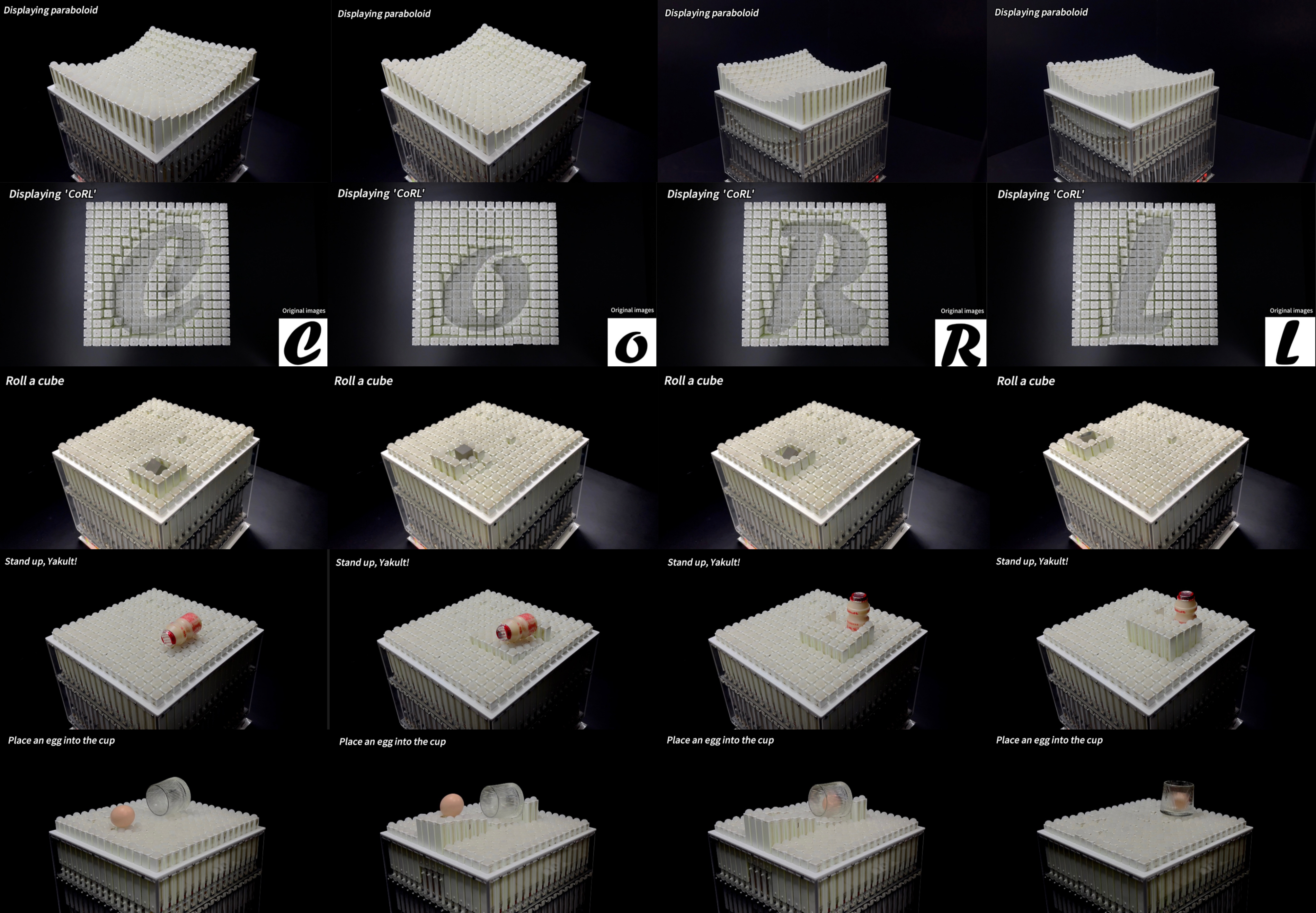}
    \caption{Applications empowered by hard-coding}
    \label{fig:app-hardcode}
\end{figure}

We show more tasks that are developed using RL on \ours in Figure~\ref{fig:app-RL}. We observe that our method can generalize well to various objects, even under heavy human perturbation or visual disturbance. Additionally, we also achieve several intriguing applications by using pre-coded motion primitives on \ours, as shown in Figure~\ref{fig:app-hardcode}. The achieved behaviors show the potential of \ours to accomplish diverse and harder tasks. 

\end{document}